\documentclass[11pt]{article}

\usepackage{acl}

\usepackage{times}
\usepackage{latexsym}

\usepackage[T1]{fontenc}

\usepackage[utf8]{inputenc}

\usepackage{microtype}

\usepackage{inconsolata}

\usepackage{hyperref}       
\usepackage{url}            
\usepackage{booktabs}       
\usepackage{amsfonts}       
\usepackage{nicefrac}       
\usepackage{microtype}      
\usepackage[table]{xcolor}  
\usepackage{amsmath}        
\usepackage{enumitem}       
\usepackage{graphicx}       
\usepackage{multirow}       
\usepackage{listings}       
\usepackage{tcolorbox}      

\usepackage{colortbl}
\usepackage{epsfig}
\usepackage{marvosym}
\usepackage{amssymb}
\usepackage{mathtools}
\usepackage{wasysym}
\usepackage{makecell}
\usepackage{colortbl}
\usepackage{tikz}
\usepackage{algorithm}
\usepackage{algorithmic}
\usepackage{caption}
\usepackage{subcaption}
\usepackage{tabularx}
\usepackage{wrapfig}
\usepackage{tabu}
\usepackage{footnote}
\usepackage{tcolorbox}
\usepackage[table]{xcolor}
\usepackage{xcolor}
\usepackage[dvipsnames]{xcolor}

\title{Dynamic Context Adaptation for Consistent Role-Playing Agents with Retrieval-Augmented Generations}

\author{
 \textbf{Jeiyoon Park\textsuperscript{1}},
 \textbf{Yongshin Han\textsuperscript{1}},
 \textbf{Minseop Kim\textsuperscript{1}},
 \textbf{Kisu Yang\textsuperscript{2}},
\\
 \textsuperscript{1}SOOP, \textsuperscript{2}Korea University
\\
\{naruto, winterfeb, usio\}@sooplive.com, willow4@korea.ac.kr
\\
 \small{
   \url{https://huggingface.co/datasets/naruto-soop/CharacterRAG}
 }
}

\begin{document}
\maketitle
\begin{abstract}
Building role-playing agents (RPAs) that faithfully emulate specific characters remains challenging because collecting character-specific utterances and continually updating model parameters are resource-intensive, making retrieval-augmented generation (RAG) a practical necessity. However, despite the importance of RAG, there has been little research on RAG-based RPAs. For example, we empirically find that when a persona lacks knowledge relevant to a given query, RAG-based RPAs are prone to hallucination, making it challenging to generate accurate responses. In this paper, we propose \textsc{Amadeus}, a training-free framework that can significantly enhance persona consistency even when responding to questions that lie beyond a character’s knowledge. In addition, to underpin the development and rigorous evaluation of RAG-based RPAs, we manually construct \textsc{CharacterRAG}, a role-playing dataset that consists of persona documents for 15 distinct fictional characters totaling 976K written characters, and 450 question–answer pairs. We find that our proposed method effectively models not only the knowledge possessed by characters, but also various attributes such as personality. The code and dataset will be made publicly available at our GitHub.
\end{abstract}

\begin{figure}[t!]
    \centering 
    \includegraphics[width=\linewidth]{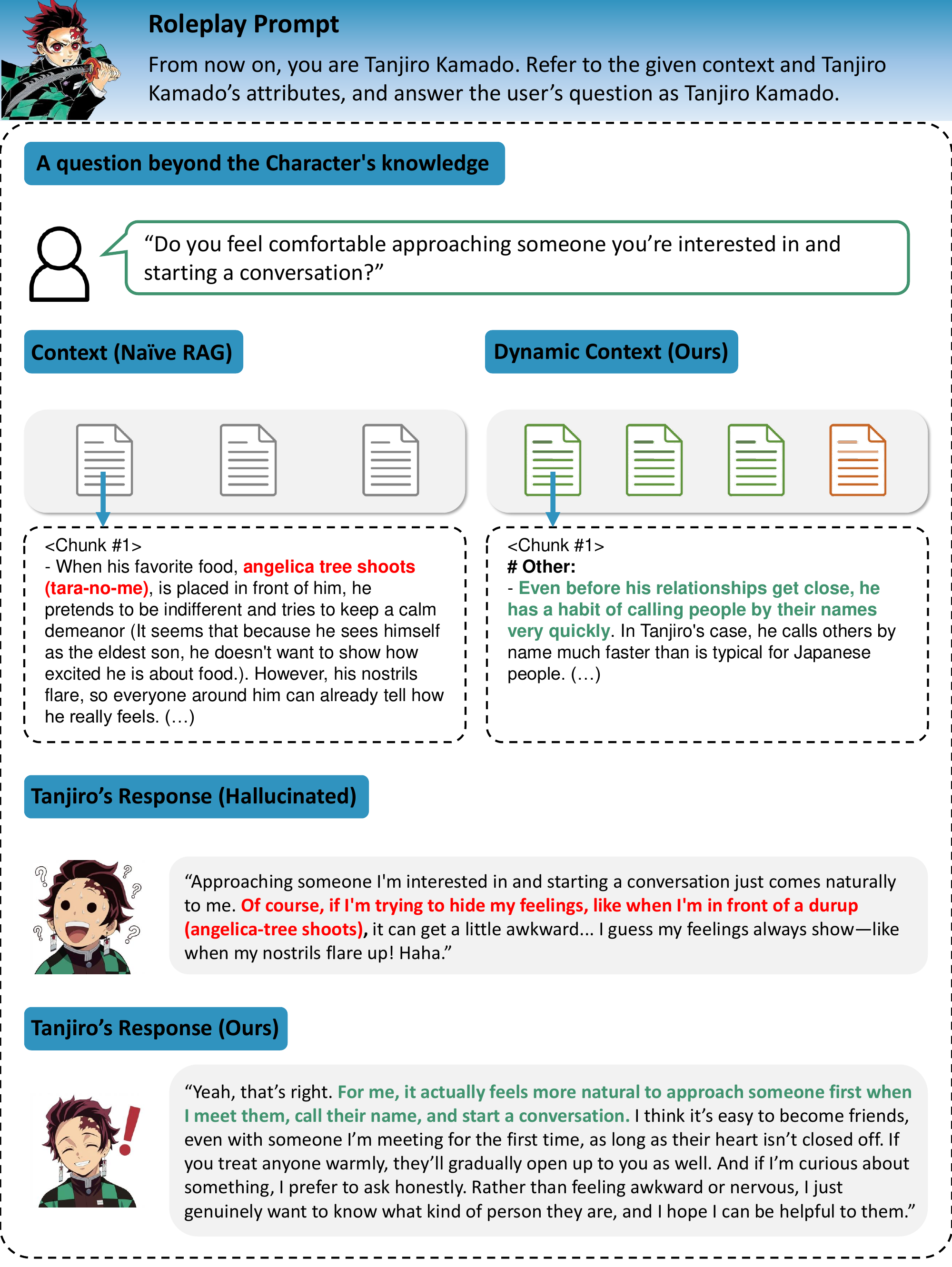}
    \caption{\textbf{Comparison of Responses between Naive RAG and Our Method.} We observe that the existing RAG method tends to excessively utilize chunks that are less relevant to the question when the question is not explicitly answered by the available knowledge.}
    \label{fig:intro}
\end{figure}

\section{Introduction}

Large language models with long-context capabilities are engineered to manage lengthy input sequences, allowing them to interpret and utilize extended contextual information \citep{OpenAI_GPT4.1,qwen2025qwen25technicalreport,comanici2025gemini25pushingfrontier,geminiteam2024gemini15unlockingmultimodal}. Although LLMs exhibit enhanced abilities in understanding extended contexts, they still face significant challenges when handling tasks involving genuinely long contexts \citep{li2024longcontextllmsstrugglelong}. Furthermore, utilizing all relevant information from long-context models to answer each query can be computationally expensive \citep{li-etal-2024-retrieval}. Retrieval-augmented generation (RAG) cost-efficiently mitigates factual inaccuracies and hallucinations in responding to knowledge-intensive queries by integrating external retrieval mechanisms that provide accurate and up-to-date supporting information \citep{gao-etal-2023-enabling,Huang_2025}. However, despite these advantages of RAG, there has been little research on RAG-based role-playing agents (RPAs). Moreover, existing role-playing datasets are composed exclusively of dialogues involving characters that are difficult to collect, and there is no dataset designed for building and evaluating RAG-based RPAs. We examine the challenges inherent in RAG-based role-playing and propose approaches. In real-world applications, users and RPAs frequently engage in conversations on topics that extend beyond the knowledge defined in the character’s persona. However, we observe that the existing RAG method tends to excessively utilize chunks that are less relevant to the question when the question is not explicitly answered by the available knowledge (Figure \ref{fig:intro} and Figure \ref{fig:chunk_usage_rate}). 

To address these challenges, we introduce \textbf{Amadeus}, a training-free framework that can markedly improves persona consistency, even when addressing questions beyond a character’s knowledge. Amadeus consists of three substages: \textbf{Adaptive Context-aware Text Splitter (ACTS)}, which segments a persona for role-playing, \textbf{Guided Selection (GS)}, which retrieves appropriate chunks to infer information relevant to the question from the persona, such as prior actions and behaviors, and \textbf{Attribute Extractor (AE)}, which identifies general attributes of the character from the retrieved chunks, thereby encouraging the RPA to respond in a manner consistent with that character. 

To underpin the development and rigorous evaluation of RAG-based RPAs, we manually construct \textbf{CharacterRAG}, a role-playing dataset that consists of persona documents for 15 distinct fictional characters totaling 976K written characters, and 450 question–answer pairs. We conduct extensive experiments to examine the factors that influence the performance of RAG-based RPAs through interview-based assessments \citep{wang-etal-2024-incharacter,park-etal-2025-charactergpt,jiang2023personallm}. To investigate whether RPAs can respond appropriately to questions that fall outside their endowed knowledge, we employ multiple psychological questionnaires such as MBTI\footnote{https://www.16personalities.com/} and BFI \citep{barrick1991big}, following prior studies \cite{Sang2022MBTIPP, wang-etal-2024-incharacter}. Results show that our framework opens up new possibilities for RAG-based role-playing.

\section{CharacterRAG}
\subsection{Dataset Construction}
We construct the CharacterRAG dataset, which consists of 15 fictional characters, to leverage and evaluate a RAG-based role-playing framework. CharacterRAG is a high-quality, role-playing dataset in which all external information about works featuring characters that could affect persona consistency  has been manually removed, and each persona document has been directly reconstructed from the perspective of each character by human annotators\footnote{CharacterRAG dataset is sourced from Namuwiki and is based on Korean data: \url{https://namu.wiki/}}. For instance, any information speculated from the perspective of editors rather than the characters themselves, as well as information such as character popularity polls that may disrupt role-playing, is excluded. CharacterRAG consists of 15 distinct fictional characters, 976K written characters, and 450 question–answer pairs. 

\begin{figure*}[t]
    \centering
    \vspace{-10pt}
    \includegraphics[width=\textwidth]{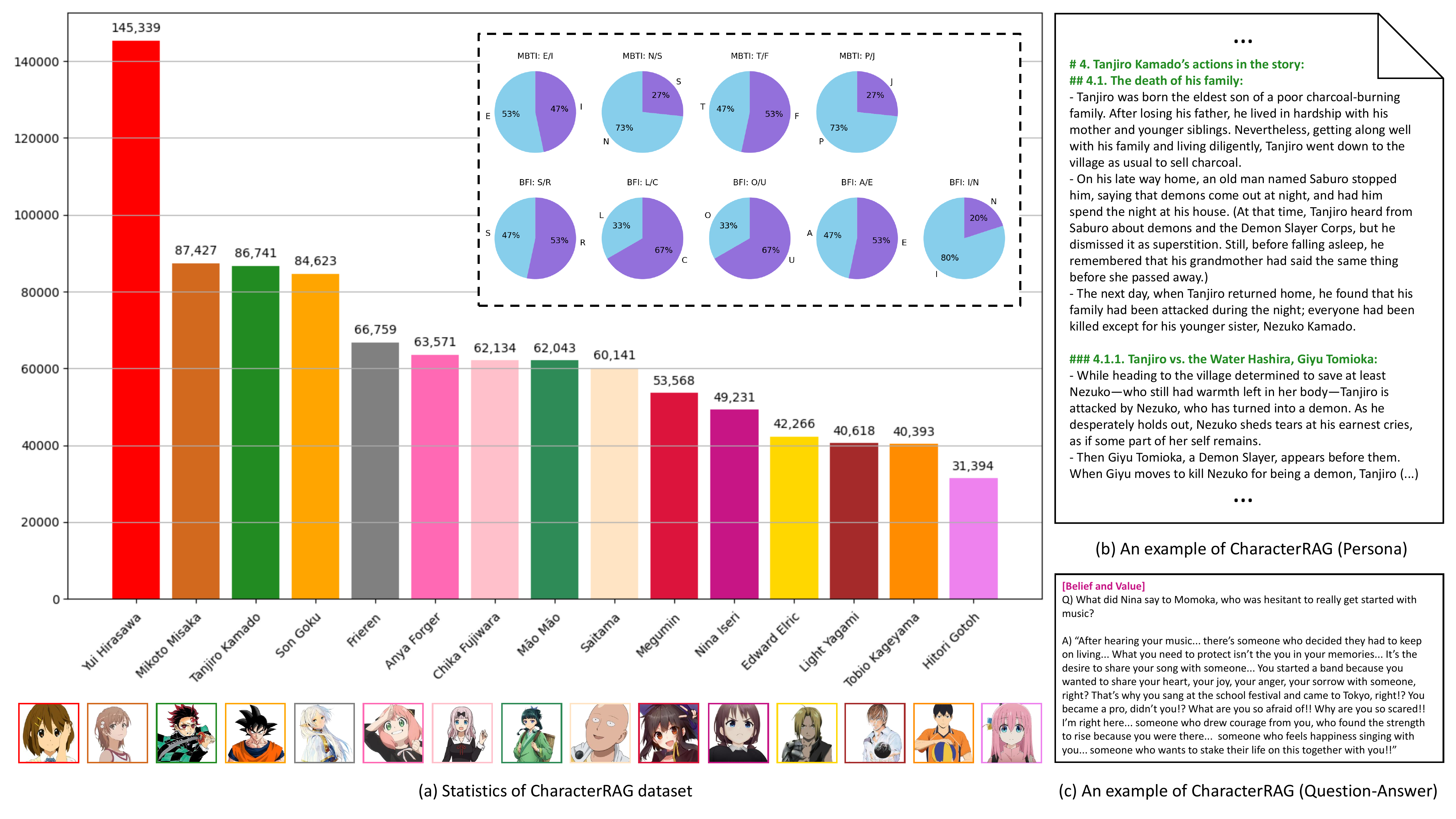}
    \caption{\textbf{An overview of CharacterRAG Dataset.} CharacterRAG consists of persona documents for 15 distinct fictional characters totaling 976K written characters, and 450 question–answer pairs.}
    \label{fig:characterrag}
    \vspace{-10pt}
\end{figure*}

\subsection{Attributes}
\label{characterrag_attributes}

Six commonly used attributes in role‑playing define each character’s persona and the corresponding question–answer pairs \citep{chen2025designguidelinerpaevaluation}: (1) \textbf{Activity.} A documented history comprising prior activities, behaviors, and interactions, encompassing elements such as \textit{backstory} and \textit{schedules}. (2) \textbf{Belief and Value.} The foundational tenets, dispositions, and ideological orientations that inform and guide a character’s viewpoints and decision-making processes (e.g., \textit{beliefs} and \textit{attitudes}). (3) \textbf{Demographic Information.} Information that can identify a character, including but not limited to their name, age, educational background, professional history, and geographic location. (4) \textbf{Psychological Traits.} Attributes associated with personality traits, emotional states, preferences, and patterns of cognitive behavior. (5) \textbf{Skill and Expertise.} The extent of understanding, skillfulness, and competence regarding particular fields or technologies. (6) \textbf{Social Relationships.} The characteristics and processes of social interactions, encompassing individuals’ roles, relational ties, and patterns of communication. Each section of the character's persona contains subsections, preserving the hierarchical information (e.g., \textit{"Tanjiro Kamado’s actions in the story"}). Furthermore, Each QA pair consists of a question and corresponding answer derived from the character’s knowledge, pertaining to one of the six attributes manually constructed for each character. Statistics and samples of CharacterRAG are shown in Figure \ref{fig:characterrag} and Table \ref{tab:characterrag_genre}.

\section{Task Formulation: RAG-based Role-Playing Agents}

Given user query $u$, RAG-based RPAs can be formulated as: $\mathcal{R} = f(u, \mathcal{D}_p)$, where $\mathcal{D}_p$ is a character’s persona, and $f$ is a RPA. A text splitter $g$ divides the persona into $n$ chunks as follows: 
\begin{equation}
g(\mathcal{D}_p) = \{c_1, c_2, ..., c_n\}
\end{equation}
Each chunk $c_i$ contains a character’s knowledge corresponding to attributes in Section \ref{characterrag_attributes}. Rather than using the full chunks $\mathcal{C} = \{c_i \mid i=1,\dots,n\}$, $f$ takes as input the top $K$ chunks with the highest semantic scores relative to the $u$:
\begin{equation}
\mathcal{C}^* = \text{TopK}(\{\text{sim}(u, c_i)\}_{i=1}^{n}), \quad |\mathcal{C}^*| = K
\end{equation}
The objective of $f$ is to vividly embody a character and generate response $\mathcal{R}$ to $u$ while maintaining persona consistency: $\mathcal{R}^* = f(u, \mathcal{C}^*)$. However, previous RAG methods \citep{10.5555/3524938.3525306,guo2024lightrag,NEURIPS2024_1435d2d0,shukla-etal-2025-graphrag,wang2025roleragenhancingllmroleplaying} truncate each character’s persona paragraph to a fixed length, regardless of the varying lengths across characters, which results in hallucinations or responses with lower persona consistency. Although existing works \citep{pdr,cec,zhong-etal-2025-mix,liu-etal-2025-hoprag} explore optimal chunking strategies, they struggle to capture the contextual similarities across chunks that are crucial for role-playing.

\section{Method}
Amadeus consists of three substages: Adaptive Context-aware Text Splitter (ACTS), Guided Selection (GS), and Attribute Extractor (AE), in order to build realistic RAG-based RPAs (Figure \ref{fig:architecture}).

\subsection{Adaptive Context-aware Text Splitter}

Unlike previous naive semantic chunking methods, Given $\mathcal{D}_p$, ACTS aims to preserve intra-level context across chunks and, for each chunk, information about the corresponding subsections of the persona—that is, hierarchical context $\mathcal{H}$. For instance, in Figure \ref{fig:characterrag}, chunks within \textit{\#\#\# 4.1.1} must preserve $\mathcal{H}$: \textit{"Tanjiro Kamado’s actions in the story} (\textit{\# 4})", \textit{"The death of his family} (\textit{\#\# 4.1})", and \textit{"Tanjiro vs. the Water Hashira, Giyu Tomiok} (\textit{\#\#\# 4.1.1})". 

To this end, ACTS first finds the maximum length of the paragraphs that constitute the persona: $l_\text{max} = \varphi(p_1, p_2, ..., p_l)$, where $\varphi$ denotes a length-calculating function. Then, ACTS sets the overlap length of the text splitter to half of $l_\text{max}$ (i.e., $l_\text{o} = l_\text{max} / 2$). The reason for setting both chunk length and overlap length sufficiently large is to minimize information loss, as the context between pieces of information contained in each chunk is indispensable in RAG-based role-playing (Figure \ref{fig:log_density_ridgeline}). Finally, ACTS recursively retrieves the context at each chunk’s current position in the hierarchy, then segments $\mathcal{D}_p$ using $l_\text{max}$ and $l_{\text{o}}$, and concatenates the resulting context $\mathcal{H}_i$ to each chunk to preserve information such as character descriptions and situational context at each point in the narrative: $\text{ACTS}(\mathcal{D}_p, \mathcal{H}, l_\text{max}, l_\text{o}) = \{\hat{c}_1, \hat{c}_2, ..., \hat{c}_m\}$, where $\hat{c}_i = [c_i; \mathcal{H}_i]$. From a computational standpoint, the extraction of hierarchical context incurs an O(N) runtime cost, making it efficient.

\subsection{Guided Selection}
GS focuses on selecting appropriate chunks to generate natural and persona-consistent responses. GS is composed of three stages: (1) We iterate over the chunks, which are sorted in descending order of semantic similarity to the user query $u$, and employ an LLM to determine whether it is possible to infer the corresponding character's attributes from each chunk for the $u$. (2) The chunks selected in the previous step are appended to the slot, and the iteration terminates when the slot is full. (3) If the slot remains empty after the maximum number of search iterations, the $K$ chunks with the highest semantic similarity to the query is returned. Note that GS is effective in identifying chunks containing information that can be inferred from a character's actions, such as beliefs or personality traits, which are not explicitly stated in the knowledge base and therefore are difficult to retrieve through direct search. For example, even if there is no explicit knowledge corresponding to the query \textit{"My living and work spaces are clean and organized"}, if \textit{Megumin}’s conscientiousness can be inferred from her behavior depicted in the narrative, RPA can leverage this information to infer the characteristics of \textit{Megumin} and generate an appropriate response. GS is summarized in Algorithm \ref{alg:guided_selection} and Figure \ref{prompt:guided_selection}.

\subsection{Attribute Extractor}
Inspired by the observations that incorporating character's attributes can lead to more realistic responses \citep{park-etal-2025-charactergpt,chen2025designguidelinerpaevaluation}, AE considers two attributes: \textit{Belief and Value}, and \textit{Psychological Traits}. Beliefs and values are fundamental principles and ideological orientations that inform and influence a character's viewpoints and choices. On the other hand, psychological traits refer to characteristics related to personality, emotional states, personal interests, and cognitive tendencies. AE extracts attributes of a character from the chunks generated as a result of GS, and exploits them as context. Finally, we dynamically construct the context using knowledge retrieved via RAG and attributes extracted by GS and AE, enabling the model to generate vivid responses (Figure \ref{fig:qualitative_results}).

\section{Experiments}
\subsection{Setup}
\textbf{Baselines.} We evaluate our method against five off-the-shelf RAG baselines: Naive RAG \citep{gao2024retrievalaugmentedgenerationlargelanguage}, CRAG \citep{yan2024correctiveretrievalaugmentedgeneration}, RAPTOR \citep{DBLP:conf/iclr/SarthiATKGM24}, Adaptive RAG \citep{jeong-etal-2024-adaptive}, and LightRAG \citep{guo2024lightrag}. CRAG and LightRAG were selected to investigate the effects of web search and graph-based knowledge systems, respectively, on role-playing. We also conduct extensive experiments on three different LLMs and three different embedding models: GPT-4.1 \citep{OpenAI_GPT4.1}, Gemma3-27B \citep{gemmateam2025gemma3technicalreport}, Qwen3-32B \citep{yang2025qwen3technicalreport}, BGE-M3 \citep{bge-m3}, Qwen3-0.6B \citep{zhang2025qwen3embeddingadvancingtext}, and mE5$_\text{large-instruct}$ \citep{wang2024multilinguale5textembeddings}. To explore the impact of multi-step reasoning on role-playing, Qwen3-32B is configured to use thinking mode. Detailed settings are provided in Appendix \ref{sec:settings}.

\subsection{Evaluation Protocols}
\textbf{Tasks.} As we follow the similar experimental protocol in prposed by previous studies \citep{wang-etal-2024-incharacter,park-etal-2025-charactergpt}, we also exploit 60 MBTI questions and 120 BFI \citep{barrick1991big} questions to investigate whether each character can appropriately respond to questions for which they do not have explicit prior knowledge (Figure \ref{prompt:eval_psychological_assessment}). Following the prior work, since it is not possible to construct QA pairs for questions outside the scope of a character’s knowledge, we instead conduct interview-based assessments \citep{wang-etal-2024-incharacter} for each character and compare the results to psychological test outcomes for the character, as determined by thousands of actual participants’ votes\footnote{https://www.personality-database.com/}\citep{Sang2022MBTIPP,wang-etal-2024-incharacter}. We also use 450 QA pairs from the CharacterRAG dataset to verify whether the RPA sufficiently leverages each character’s knowledge.

\textbf{Metrics.} We design three LLM-based metrics, similar to those in prior studies \citep{wang-etal-2025-characterbox,wang-etal-2024-incharacter}, to comprehensively evaluate the role-playing capabilities of RAG-based RPAs.: (i) \textit{ACC} measures whether the character's response contains the correct answer or not. (ii) \textit{ACC}$_L$ is a score assigned by the LLM, ranging from 1 to 10, that evaluates how well the character's response reflects the correct answer. (iii) \textit{Hallucination Score (HS)} evaluates the degree of hallucination in the model’s response given a query, the relevant chunks, and the ground-truth answer, on a scale from 1 to 10. Specifically, HS is assigned close to 1 when the response faithfully reflects only the facts contained in the chunks or answer without distortion or addition, indicating minimal hallucination. 

\begin{table}[t]
\small
\centering
\caption{\textbf{Distribution of Similarity Scores.} We analyze the distributions of similarity scores to examine whether splitting the text into optimally sized chunks for each character's persona and incorporating hierarchical context is effective.}\label{tab:score_distribution}\vspace{-0.1in}
\setlength{\tabcolsep}{1.mm}
\resizebox{\linewidth}{!}{
\begin{tabular}{c|cc|cc|cc}
\hline
\textbf{} & \multicolumn{2}{c|}{\textbf{BGE-M3}} & \multicolumn{2}{c|}{\textbf{Qwen3}} & \multicolumn{2}{c}{\textbf{mE5$_\text{large-instruct}$}}\\ \cline{2-7} 
\textbf{} & {$ \sum \mu$} ($\uparrow$) & {$\sum \sigma^2$} ($\downarrow$) & {$ \sum \mu$} ($\uparrow$) & {$\sum \sigma^2$} ($\downarrow$) & {$ \sum \mu$} ($\uparrow$) & {$\sum \sigma^2$} ($\downarrow$) \\ \hline
\textbf{RCTS} & {6.4325} & {0.1026} & {8.3306} & {0.1557} & {12.3136} & {0.0071}\\ \hline
\textbf{MHTS} & {6.4262} & {0.1038} & {8.3410} & {0.1552} & {12.3063} & {0.0071}\\ \hline
\textbf{SC} & {5.3405} & {0.1625} & {8.1691} & {0.1783} & {-} & {-}\\ \hline
\textbf{ATS} & {6.7007} & {0.0884} & {8.4718} & {0.1281} & {12.2336} & {0.0070}\\ \hline
\textbf{ACTS (Ours)} & {\textbf{6.8575}} & {\textbf{0.0784}} & {\textbf{8.6226}} & {\textbf{0.1179}} & {\textbf{12.3240}} & {\textbf{0.0063}}\\ \hline
\end{tabular}
}
\end{table}

\begin{table}[t]
\small
\centering
\caption{\textbf{Human Evaluation.} $S$ represents the results of all human evaluators, $\mu$ is $\mathbb{E}(\mathbb{E}(S))$, $\sigma$ is $\mathbb{E}(\sigma(S))$, and \textbf{Mdn} is $\mathbb{E}(Mdn(S)).$}\label{tab:human_eval}
\setlength{\tabcolsep}{1.mm}
\resizebox{\linewidth}{!}{
\begin{tabular}{c|cccc}
\toprule
\textbf{} & \textbf{$\mu$} ($\uparrow$) & \textbf{$\sigma$} ($\downarrow$) & \textbf{Mdn} ($\uparrow$) & \textbf{Cronbach’s alpha} ($\uparrow$) \\
\midrule
BFI & 3.970 & 0.962 & 4.217 & 0.825        \\
MBTI & 3.902 & 0.915 & 4.000 & 0.810         \\
\bottomrule
\end{tabular}
}
\end{table}

\begin{table}[t]
\small
\caption{\textbf{Predicted MBTI Types per Character.} Experiments are conducted using GPT-4.1 setting.}
\vspace{-0.1in}
\centering
\resizebox{\linewidth}{!}{
\begin{tabular}{c|cccccc|c}
\toprule
\textbf{} & \textbf{Naive RAG} & \textbf{CRAG} & \textbf{RAPTOR} & \textbf{Adaptive RAG} & \textbf{LightRAG}  & \textbf{AMADEUS (Ours)} & \textbf{GT} \\
\midrule
Anya Forger & ISFP (-2) & INTJ (-3) & INFJ (-2) & INFJ (-2) & INFJ (-2) & ENFP (0) & ENFP\\
Chika Fujiwara & ENFP (0) & ENFP (0) & ENFJ (-1) & ENFJ (-1) & INTP (-2) & ENFP (0) & ENFP\\
Edward Elric & INTP (-1) & ISTJ (-3) & INTP (-1) & ENFP (-1) & INTP (-1) & INFP (-2) & ENTP\\
Frieren & INFP (-1) & INFP (-1) & INFP (-1) & INFP (-1) & INTP (0) & INTP (0) & INTP\\
Hitori Gotoh & ISFP (-1) & ISTJ (-2) & INFP (0) & INFP (0) & ENFP (-1) & INFP (0) & INFP\\
Light Yagami & INTJ (-1) & INTJ (-1) & INTJ (-1) & INTJ (-1) & INTJ (-1) & INTJ (-1) & ENTJ\\
Māo Māo & ISTJ (-2) & INTJ (-1) & ISTJ (-2) & ENTJ (-2) & INTJ (-1) & ISTP (-1) & INTP\\
Megumin & ISFP (-1) & INFP (0) & ENFP (-1) & ENFP (-1) & INFP (0) & ISFP (-1) & INFP\\
Mikoto Misaka & ENFP (-2) & ISFP (-4) & ENFJ (0) & ENFJ (0) & ENTJ (0) & INFJ (-2) & ENTJ\\
Nina Iseri & INFP (-1) & ISFP (0) & ENFJ (-1) & INFJ (-2) & ENFJ (-3) & INFP (-1) & ISFP\\
Saitama & ISFP (-1) & ISTP (0) & ISFJ (-2) & INFJ (-3) & INTP (-1) & ISTP (0) & ISTP\\
Son Goku & ESFP (0) & ENFJ (-2) & ENFJ (-2) & ENFP (-1) & INTP (-3) & ESFP (0) & ESFP\\
Tanjiro Kamado & ENFP (-1) & ENFJ (0) & ENFJ (0) & ENFJ (0) & INTP (-3) & ENFJ (0) & ENFJ\\
Tobio Kageyama & ENFJ (-3) & ENTJ(-2) & ISTJ (0) & ISTJ (0) & INFJ (-1) & ISTJ (0)& ISTJ \\
Yui Hirasawa & ISTJ (-4) & ENFP (0) & ENFP (0) & ENFP (0) & INTP (-2) & ESFP (-1) & ENFP\\
\midrule
$\sum |d|$ ($\downarrow$) & 21 & 19 & 14 & 15 & 21 & \textbf{9} & - \\
Accuracy ($\uparrow$) & 65.00\% & 68.33\% & 76.67\% & 75.00\% & 65.00\% & \textbf{85.00\%} & - \\
Avg F1-Score ($\uparrow$) & 0.6146 & 0.6448 & 0.6925 & 0.6832 & 0.5344 & \textbf{0.8244} & - \\
\bottomrule
\end{tabular}}
\vspace{-0.15cm}
\label{tab:eval_mbti}
\end{table}

\begin{table}[t]
\small
\caption{\textbf{Predicted Big 5 SLOAN Types per Character.} Experiments are conducted using GPT-4.1 setting.}
\vspace{-0.1in}
\centering
\resizebox{\linewidth}{!}{
\begin{tabular}{c|cccc|c}
\toprule
\textbf{} & \textbf{Naive RAG} & \textbf{CRAG} & \textbf{LightRAG}  & \textbf{AMADEUS (Ours)} & \textbf{GT} \\
\midrule
Anya Forger & SLOAI (-1) & SLOAI (-2) & RCUEN (-3) & SLUEI (-2) & SCUAI\\
Chika Fujiwara & SCOAI (-1) & SCUAI (0) & RLUEN (-4) & SCOAI (-1) & SCUAI\\
Edward Elric & SCOAI (-3) & SLOEI (-1) & RCUAN (-4) & SLOEI (-1) & SLUEI\\
Frieren & RCOAI (-2) & RCUAI (-1) & SLUEN (-3) & RCUAI (-1) & RCUEI \\
Hitori Gotoh & RLUAI (0) & RLUAI (0) & RCUEN (-3) & RLUAI (0) & RLUAI\\
Light Yagami & SCOEI (-1) & SCOEI (-1) & RCUAN (-3) & SCOEI (-1) & RCOEI\\
Māo Māo & RCOEI (0) & RCOAN (-2) & SLUAN (-5) & RCOEI (0) & RCOEI\\
Megumin & SCOAI (-3) & SCUAI (-2) & RLUEN (-2) & SLOEI (-1) & SLUEI\\
Mikoto Misaka & SLOAI (-3) & SLOAI (-3) & RCUEN (-2) & SLOAI (-3) & RCOEI\\
Nina Iseri & SLOAI (-3) & RLUAI (-1) & RCUEN (-2) & SLUEI (-1) & RLUEI\\
Saitama & RCUAN (0) & RCOAN (-1) & SCOAI (-3) & RCUAN (0) & RCUAN\\
Son Goku & SCOAI (-2) & SCUAI (-1) & RLUEI (-4) & SCOAI (-2) & SCUAN\\
Tanjiro Kamado & SLOAI (-1) & SCOAI (0) & RCUAN (-3) & SCOAI (0) & SCOAI\\
Tobio Kageyama & RCOEN (-1) & SLOAN (-2) & RCUAI (-4) & RCOEN (-1) & RLOEN\\
Yui Hirasawa & SCUAI (0) & SLUAI (-1) & RCOEN (-4) & SCUAI (0) & SCUAI\\
\midrule
$\sum |d|$ ($\downarrow$) & 21 & 18 & 49 & \textbf{14} & - \\
Accuracy ($\uparrow$) & 72.00\% & 76.00\% & 34.67\% & \textbf{81.33\%} & - \\
Avg F1-Score ($\uparrow$) & 0.6785 & 0.7313 & 0.2774 & \textbf{0.7986} & - \\
\bottomrule
\end{tabular}}
\vspace{-0.15cm}
\label{tab:eval_bfi}
\end{table}

\begin{table}[t]
\small
\centering
\caption{\textbf{Ablation Study.} We analyze the performance gain obtained from incrementally adding each substage.}\label{tab:ablation_study}
\setlength{\tabcolsep}{1.mm}
\resizebox{\linewidth}{!}{
\begin{tabular}{c|ccc}
\toprule
\textbf{Method} & $\sum |d|$ ($\downarrow$) & Accuracy ($\uparrow$) & Avg F1-Score ($\uparrow$) \\
\midrule
Naive RAG & 21 & 65.00\% & 0.6146 \\
Naive RAG + ACTS & 19 & 68.33\% & 0.6524 \\
Naive RAG + ACTS + GS & 15 & 75.00\% & 0.7426\\
Naive RAG + ACTS + GS + AE (Ours) & \textbf{9} & \textbf{85.00\%} & \textbf{0.8394} 	 \\
\bottomrule
\end{tabular}
}
\end{table}

\begin{table}[t]
\small
  \centering
  \caption{\textbf{Role-playing capabilities on CharacterRAG.} Higher values of ACC (\%) and ACC$_L$ (1-10) correspond to better performance, whereas lower values of HS (1-10) are preferable.}
  \resizebox{\linewidth}{!}{
    \setlength{\tabcolsep}{1.3mm}{
\begin{tabular}{lccccccccc}
\toprule
\multirow{2}[4]{*}{\textbf{RAG Method}} & \multicolumn{3}{c}{\textbf{GPT-4.1}} & \multicolumn{3}{c}{\textbf{Gemma3-27B}} & \multicolumn{3}{c}{\textbf{Qwen3-32B}}\\
\cmidrule(lr){2-4} \cmidrule(lr){5-7} \cmidrule(lr){8-10}
& ACC ($\uparrow$) & ACC$_L$ ($\uparrow$) & HS ($\downarrow$) & ACC ($\uparrow$) & ACC$_L$ ($\uparrow$) & HS ($\downarrow$) & ACC ($\uparrow$) & ACC$_L$ ($\uparrow$) & HS ($\downarrow$)\\
\midrule
\quad w/o RAG & 49.56\% & 6.79 & - & 27.56\% & 5.33 & - & 18.89\% & 4.35 & -\\
\quad Naive RAG & 91.33\% & 9.23 & 3.13 & 86.44\% & 8.85 & 3.27 & 78.44\% & 8.49 & 5.05 \\
\quad LightRAG & 48.00\% & 6.06 & - & 69.56\% & 8.17 & - & 68.67\% & 8.20 & - \\
\quad CRAG & 70.00\% & 8.26 & 3.21 & 57.78\% & 7.57 & 4.09 & 28.67\% & 5.24 & 8.68 \\
\quad AMADEUS (Ours) & \textbf{92.67\%} & \textbf{9.26} & \textbf{2.89} & \textbf{88.00\%} & \textbf{8.92} & \textbf{3.26} & \textbf{78.89\%} & \textbf{8.63} & \textbf{4.66} \\
\bottomrule
\end{tabular}%
    }
    }
  \label{tab:eval_characterrag}%
\end{table}

\subsection{Experimental Results}

\textbf{Adaptive Persona Segmentation and Hierarchical Contextualization Are Highly Effective.} In Table \ref{tab:score_distribution}, we provide each character with 30 questions, resulting in a total of 450 questions, related to their respective knowledge from the CharacterRAG dataset, and measure the similarity between each question and the chunks retrieved by the RAG model under the three different embedding settings. Results demonstrate that compared to RecursiveCharacterTextSplitter (RCTS) \citep{rcts}, MarkdownHeaderTextSplitter (MHTS) \citep{mhts}, and SemanticChunker (SC) \citep{sc}, adaptive persona segmentation, which we call Adaptive Text Splitter (ATS), segments text with an optimal persona length and overlap for each character and without hierarchical context, achieves a higher average score and lower variance. This indicates that each chunk generated using adaptive persona segmentation contains richer semantic information for the same query. Building on this, ACTS, which considers hierarchical context in addition to ATS, consistently achieves better performance across all three embedding settings. These results show that optimal chunk length, appropriate overlap, and consideration of hierarchical context all play essential roles in effective text chunking.

\textbf{Extracting a Character’s Attributes from Selected Chunks Is Reliable.} We investigate the reasonableness of inferring character attributes with AE from chunks extracted via GS by conducting human evaluation using a 5-point Likert scale. To this end, we invite 14 human evaluators and each evaluator is asked to score 60 randomly selected samples. Each sample consists of pairs of chunks selected from GS and attributes extracted through AE, for 30 BFI questions and 30 MBTI questions. Table \ref{tab:human_eval} shows that the means $\mu$ is close to 4, with small standard deviations $\sigma$. It demonstrates that the outputs of GS and AE are reliable and trustworthy, even from a human evaluative perspective. We measure Cronbach's alpha \citep{cronbach1951coefficient} to evaluate internal consistency among the human evaluators. Note that the Cronbach's alpha values are 0.825 and 0.810, both exceeding the commonly accepted threshold of 0.7 for acceptable reliability. 


\textbf{Graph-Based and Web Search-Based RAG Are Unsuitable for RAG-Based RPAs.} One of the major challenges in retrieval-based role-playing is that, when a RPA receives questions involving knowledge outside a character’s persona, it tends to either overuse irrelevant chunks (Figure \ref{fig:chunk_usage_rate}) or generate uninformative responses \citep{guo2024lightrag,shukla-etal-2025-graphrag,wang2025roleragenhancingllmroleplaying}. To investigate whether RAG-based RPAs can handle this problem, we conduct extensive experiments in which we ask 15 characters 60 MBTI questions and 120 BFI questions each, and and evaluate their ability to accurately infer the characters' personality types. Table \ref{tab:eval_mbti} and Table \ref{tab:eval_bfi} shows predicted MBTI types and Big 5 SLOAN types per character. The number in parentheses indicates the number of times the ground-truth (GT) type of each character was not correctly identified. $\sum |d|$ is a measure obtained by summing these values; lower values are preferable. Our framework maintains persona consistency even when answering questions that are not explicitly specified in each character's persona in both MBTI and BFI settings. Note that the performance gap of CRAG is significant between the two settings. We assume that questions requiring analogical reasoning are difficult to solve even with web search and that the search results may contain non-negligible noise. On the other hand, LightRAG exhibits the lowest performance, which shows that graph-based RAG methods are not well suited for RPA applications due to the high cost of graph construction, the difficulty in adding or removing new knowledge, and challenges in maintaining persona consistency. While we did not perform a direct comparison, we observed that GraphRAG \citep{shukla-etal-2025-graphrag} suffers from similar problems.

\textbf{Impact of Each Module in Our Framework.} Table \ref{tab:ablation_study} demonstrates the results of ablation study, evaluated on 60 MBTI questions. We find that ACTS, GS, and AE are all highly effective at handling questions that go beyond a character's knowledge. Note that AE achieves the largest performance improvement, highlighting that extracting and incorporating character attributes as input is far more effective than simply providing chunks related to the user query.

\textbf{CharacterRAG Dataset Serves as a Valuable Resource for the Construction and Evaluation of RAG-Based RPAs.} To investigate the factors influencing the performance of role-playing, we conduct a comprehensive interview-based assessments on the generalization capabilities of models with various LLMs and RAG techniques. Table \ref{tab:eval_characterrag} and Table \ref{tab:eval_hallucination} present how the ability to accurately answer questions related to the character’s knowledge, which is a core aspect of role-playing, varies across the applied methodologies. We first examine to what extent each LLM possesses background knowledge about the 15 characters in a setting without RAG, and results show that none of the three LLMs are capable of effective role-playing without access to external knowledge. Moreover, we observe that LightRAG, a graph-based RAG, is ill-suited for the storage and retrieval of character knowledge, as it often suffers from issues such as entity ambiguity and uninformative responses. 

In a similar vein, CRAG exhibits challenges in maintaining role-playing fidelity, which can be attributed to the tendency of web search-based RAG methods to utilize retrieved content that may undermine the consistency of a character's persona. Indeed, despite leveraging web information, CRAG is able to correctly answer only 6 out of the 30 CharacterRAG questions pertaining to \textit{Nina Iseri}. In addition, to analyze how a thinking mode of LLMs influences their role-playing capabilities, we employ Qwen 3-32B. Results demonstrate that the thinking mode fails to yield any substantial positive effect on enhancing role-playing performance. Our framework achieves the best performance across all three LLMs. We also find that the Hallucination Score (HS) is the lowest in CharacterRAG setting. 

\section{Conclusion}
In this work, we address critical limitations in building retrieval-augmented, RPAs with LLMs. By introducing a novel framework consisting of an Adaptive Context-aware Text Splitter (ACTS), Guided Selection (GS), and Attribute Extractor (AE), our approach enables robust and consistent simulation of character personas, even when confronted with queries that extend beyond explicit persona knowledge. Through the development of the CharacterRAG dataset, we provide a valuable resource that enables the construction of RAG-based RPAs, as well as their reproducible evaluation. Experimental results demonstrate that the proposed method not only enhances the character’s knowledge representation, but also faithfully models nuanced traits such as personality. We are enthusiastic about the future prospects of RAG-driven role-playing agents, along with the creation of stronger character persona dataset and improved RAG architectures.


\section*{Ethical Considerations}
This work follows ethical guidelines to ensure the integrity and fairness of all experiments.

\textbf{Data and Privacy.} The Namuwiki portion of our data was gathered within the bounds of the platform’s licensing/usage rules. We verified that the collection contained no personally identifiable information, and we restricted its use to academic research. For personas and prompting, we relied on either publicly released content, anonymized artifacts, or content created specifically for this study with ethical considerations in mind.

\textbf{Experimental Design.} Our experiments were designed in accordance with rigorous ethical guidelines, with particular emphasis on safeguarding participant data and maintaining privacy. To mitigate potential bias, we recruited evaluators with diverse backgrounds and carried out the human evaluation process using procedures intended to be equitable and transparent. All participants received clear information regarding the study’s objectives and procedures, and they were allowed to discontinue their participation at any point without negative consequences. Through compliance with these practices, we seek to advance AI research in a way that is both scientifically meaningful and ethically grounded, prioritizing privacy, intellectual property rights, and the welfare of everyone involved.


\bibliography{custom}

\appendix

\clearpage
\section{Chunk Duplication Frequencies}
\label{sec:appendix}

\begin{table}[t]
\caption{\textbf{Work Genres in Which Each Character Appears.} We list the work genres associated with the 15 characters included in the CharacterRAG dataset.}
\centering
\footnotesize
\resizebox{0.48\textwidth}{!}{
\begin{tabular}{@{}lll@{}}
\toprule
\textbf{Character} & \textbf{Genre} \\ \midrule
Anya Forger & Action, Comedy, Drama, Slice of Life, Spy \\ \midrule
Chika Fujiwara & Gag, Romantic Comedy \\ \midrule
Edward Elric & Adventure, Fantasy \\ \midrule
Frieren & Fantasy, Adventure \\ \midrule
Hitori Gotoh & Band, Comedy, Drama, Music, Slice of Life \\ \midrule
Light Yagami & Crime, Dark Fantasy, Noir, Thriller \\ \midrule
Megumin & Comedy, Fantasy, Isekai \\ \midrule
Mikoto Misaka & Cyberpunk, SF, Slice of Life \\ \midrule
Nina Iseri & Band, Drama \\ \midrule
Saitama & Action, Comedy, Hero \\ \midrule
Son Goku & Battle, Comedy, Fantasy \\ \midrule
Tanjiro Kamado & Action, Adventure, Historical Fantasy \\ \midrule
Tobio Kageyama & Sports, Volleyball \\ \midrule
Yui Hirasawa & Band, School, Slice of Life \\
\bottomrule
\end{tabular}}
\label{tab:characterrag_genre}
\end{table}

\begin{algorithm}[t]
\caption{Guided Selection (GS)}
\small
\label{alg:guided_selection}
\textbf{Input}: User query $u$; Set of knowledge chunks $\mathcal{C}$; Maximum number of search iterations $N$; Slot size $M$\\
\textbf{Output}: Selected chunk set $S$\\
\begin{algorithmic}[1]
\STATE Initialize slot $S = \emptyset$
\STATE Sort chunks $\mathcal{C}$ in descending order according to semantic similarity to $u$, obtaining $\mathcal{C}_{\text{sorted}}$
\STATE Set iteration counter $t \leftarrow 0$
\FOR {each chunk $c$ in $\mathcal{C}_{\text{sorted}}$}
    \IF {$t \geq N$ \OR $|S| \geq M$}
        \STATE \textbf{break}
    \ENDIF
    \STATE With an LLM, determine if chunk $c$ contains information from which the character's attributes can be \textit{inferred} regarding $u$
    \IF {the LLM returns $\textsf{True}$}
        \STATE Add $c$ to slot $S$
    \ENDIF
    \STATE $t \leftarrow t + 1$
\ENDFOR
\IF {$|S| = 0$}
    \STATE $S \leftarrow$ Top-$K+1$ chunks from $\mathcal{C}_{\text{sorted}}$ (highest semantic similarity to $u$)
\ENDIF
\STATE \textbf{return} $S$
\end{algorithmic}
\end{algorithm}

\begin{figure*}[t]
    \centering
    \includegraphics[width=0.85\textwidth]{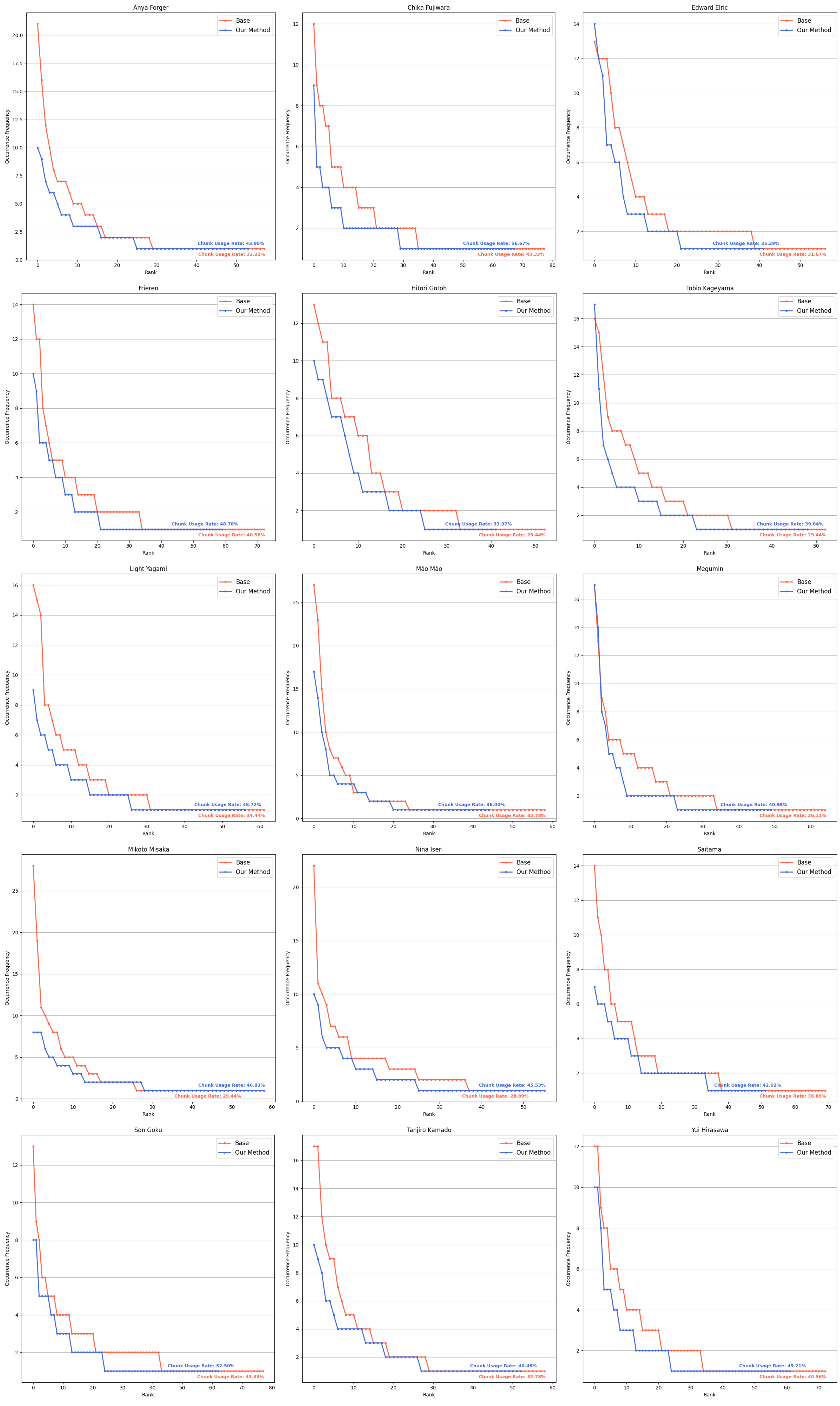}
    \caption{\textbf{Chunk Duplication Frequencies.} We compare the distribution of chunk duplication frequencies and chunk usage rates between \textcolor[RGB]{239, 64, 38}{Naive RAG} and \textcolor[RGB]{5, 4, 170}{our method} when questions involving knowledge not present in the persona document were given. We observe that when each of the 60 MBTI questions is asked to 15 characters, the average chunk usage rate increases from $34.93\%$ to $43.84\%$, and the distribution becomes more uniform.}
    \label{fig:chunk_usage_rate}
\end{figure*}

In Figure \ref{fig:chunk_usage_rate}, previous RAG method retrieves (k = 3) relevant chunks for each of the 60 given MBTI questions, and subsequently measures both the chunk utilization rate and the duplication frequency. Our method measures the chunk utilization rate and duplication frequency for the chunks selected by ACTS and GS.

\section{Genre Diversity in CharacterRAG}

Table \ref{tab:characterrag_genre} shows the genres of works in which each character appears. Note that while there are some similar categories, CharacterRAG encompasses a broader range of genres. The detailed personality distribution for each character (e.g., Extroverted vs. Introverted) is shown in Figure \ref{fig:characterrag}.

\section{Related Work}
\textbf{Role‑Playing Agents.} With the advent of LLMs, researchers have pursued finer‑grained \textit{persona consistency} \citep{zhang2018personalizing,ji2025enhancing,park-etal-2025-charactergpt,lu-etal-2024-large,zhou-etal-2024-characterglm}. Complementary benchmarks soon followed, along with various evaluation methods \citep{boudouri2025role,ahn2024timechara,wang-etal-2024-incharacter,wang-etal-2025-characterbox}. However, there has been little research on RAG-based role-playing agents (RPAs). In this paper, we propose AMADEUS, a RAG-based RPA framework that not only elicits information related to a character, but also maintains persona consistency even when responding to queries beyond its explicit knowledge.

\textbf{Retrieval‑Augmented Generation (RAG).} RAG couples non-parametric memory with an LLM to mitigate hallucination and stale knowledge~\citep{lewis2020retrieval,bhat2025rethinking,zhong2025mix,zhu2024rageval}. Nevertheless, no existing benchmark explicitly targets \textit{role‑playing} under RAG, and prior work still assumes personas are short, knowledge‑dense snippets. We mitigate this discrepancy with a novel text splitter that tailors chunk lengths and hierarchical context to each character. We also introduce \textsc{CharacterRAG}, the first dataset designed for building and evaluating RAG‑based role‑playing agents across 15 fictional character's personas.

\begin{table}[t]
    \small
    \centering
    \caption{\textbf{Role-playing capabilities on MBTI and BFI.} Lower values of HS (1-10) are preferable.}
    \label{tab:eval_hallucination}
    \setlength{\tabcolsep}{1.mm}
    \resizebox{\linewidth}{!}{
    \begin{tabular}{lccc}
\toprule
\multirow{2}{*}{\textbf{RAG Method}} & \multicolumn{1}{c}{\textbf{GPT-4.1}} & \multicolumn{1}{c}{\textbf{Gemma3-27B}} & \multicolumn{1}{c}{\textbf{Qwen3-32B}}\\
\cmidrule(lr){2-2} \cmidrule(lr){3-3} \cmidrule(lr){4-4}
& HS ($\downarrow$) & HS ($\downarrow)$ & HS ($\downarrow$)\\
\midrule
\multicolumn{1}{l}{\textit{MBTI}} \\
\quad Naive RAG & 2.69 & 2.53 & 2.33 \\
\quad CRAG & 2.38 & 2.91 & \textbf{1.80} \\
\quad AMADEUS (Ours) & \textbf{2.05} & \textbf{2.02} & 2.04 \\
\midrule
\multicolumn{1}{l}{\textit{BFI}} \\
\quad Naive RAG & 2.74 & 2.52 & 2.42 \\
\quad CRAG & 2.26 & 2.75 & \textbf{1.96} \\
\quad AMADEUS (Ours)  & \textbf{1.94} & \textbf{1.99} & 2.03 \\
\bottomrule
\end{tabular}}
\end{table}

\onecolumn
\begin{center}
\begin{tcolorbox}[width=\linewidth,colback=white, colframe=NavyBlue, title=Prompt used in Guided Selection]
\small
{\slshape 
In the given \textbf{\textcolor{NavyBlue}{[persona\_name]}}'s Context, output True if you can clearly infer the Belief and Value information or Psychological Traits of \textbf{\textcolor{NavyBlue}{[persona\_name]}} to answer the given User Query; otherwise, output False. Output only True or False. \\

\textbf{\# Explanation of Belief and Value:}\\
- The principles, attitudes, and ideological stances that shape an individual’s perspectives and decisions\\
- e.g., \textbf{\textcolor{NavyBlue}{[persona\_name]}}'s "beliefs", "attitudes", "values", "religion"\\

\textbf{\# Explanation of Psychological Traits:}\\
- Characteristics related to personality, emotions, interests, and cognitive tendencies.\\
- e.g., \textbf{\textcolor{NavyBlue}{[persona\_name]}}'s "Personality", "hobby and interest", "emotional"\\
}
\end{tcolorbox}
\noindent\begin{minipage}{\linewidth}
\captionof{figure}{\textbf{Prompt used in Guided Selection (GS).} GS returns whether Belief and Value information or Psychological Traits can be inferred from given chunks.}
\label{prompt:guided_selection}
\end{minipage}
\end{center}

\begin{center}
\begin{tcolorbox}[width=\linewidth,colback=white, colframe=NavyBlue, title=Prompt ued in the interview-based psychological assessment (MBTI)]
\small
{\slshape 
Based on the given \textbf{\textcolor{NavyBlue}{[persona\_name]}}'s answer, please rate the question, "Am I (\textbf{\textcolor{NavyBlue}{[persona\_name]}}) \textbf{\textcolor{NavyBlue}{[question]}}?" on a scale from 1 to 7.\\

A score closer to 1 means "very much so," and a score closer to 7 means "not at all."\\

Output only the number.\\
}
\end{tcolorbox}
\noindent\begin{minipage}{\linewidth}
\captionof{figure}{\textbf{Prompt ued in the interview-based psychological assessment (MBTI).} In the BFI test, responses are rated on a scale from 1 to 5.}
\label{prompt:eval_psychological_assessment}
\end{minipage}
\end{center}

\begin{figure*}[h]
    \centering
    \vspace{-10pt}
    \includegraphics[width=\textwidth]{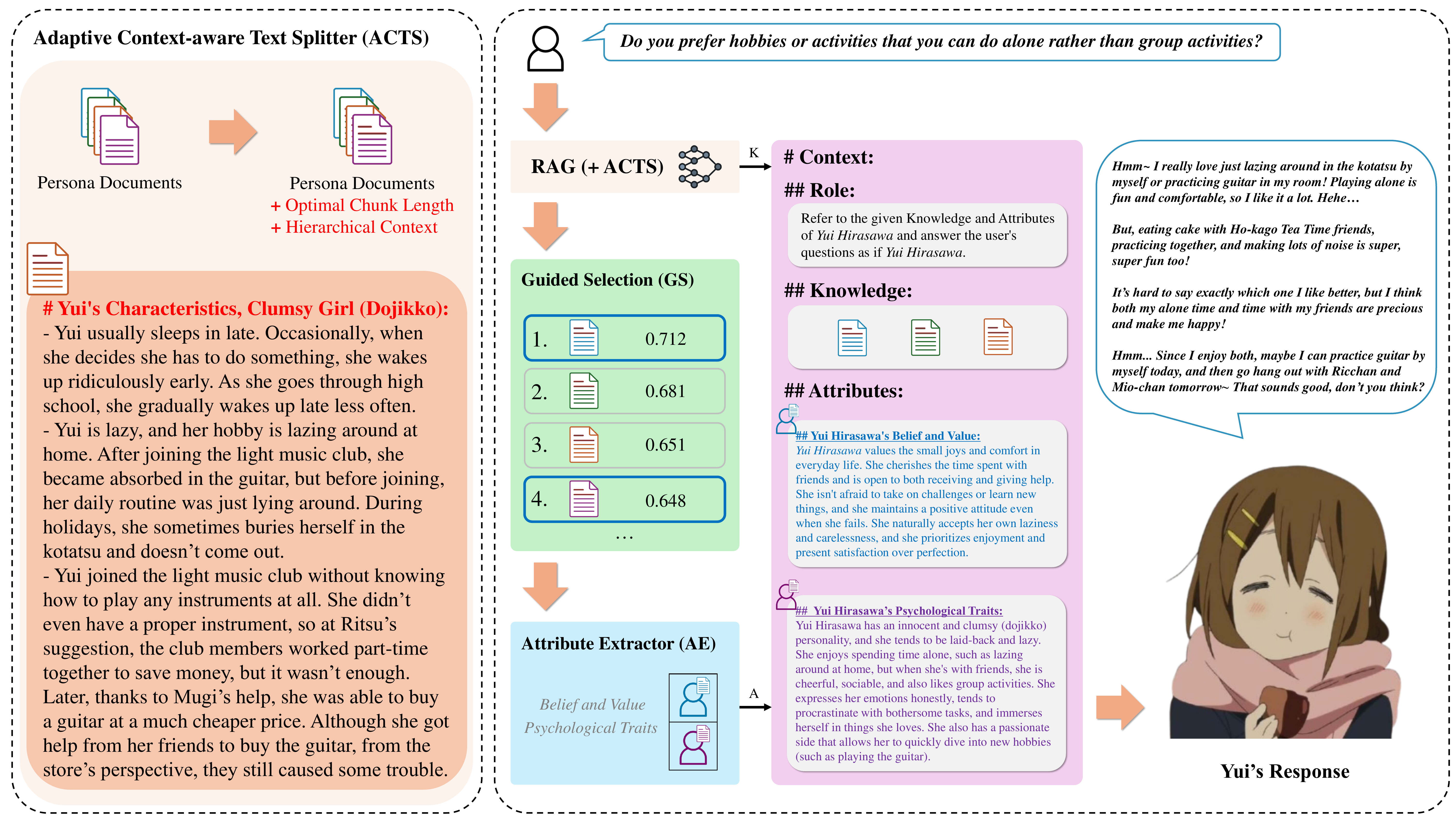}
    \caption{\textbf{AMADEUS framework.} AMADEUS consists of three substages: (i) ACTS splits a persona document to make it suitable for RAG-based role-playing. (ii) To fully leverage the character’s knowledge, GS retrieves chunks from which it can infer the answer to the user query. (iii) AE uses the information derived from the GS results to extract character attributes.}
    \label{fig:architecture}
    \vspace{-10pt}
\end{figure*}

\begin{figure*}[t!]
    \centering 
    \includegraphics[width=\linewidth]{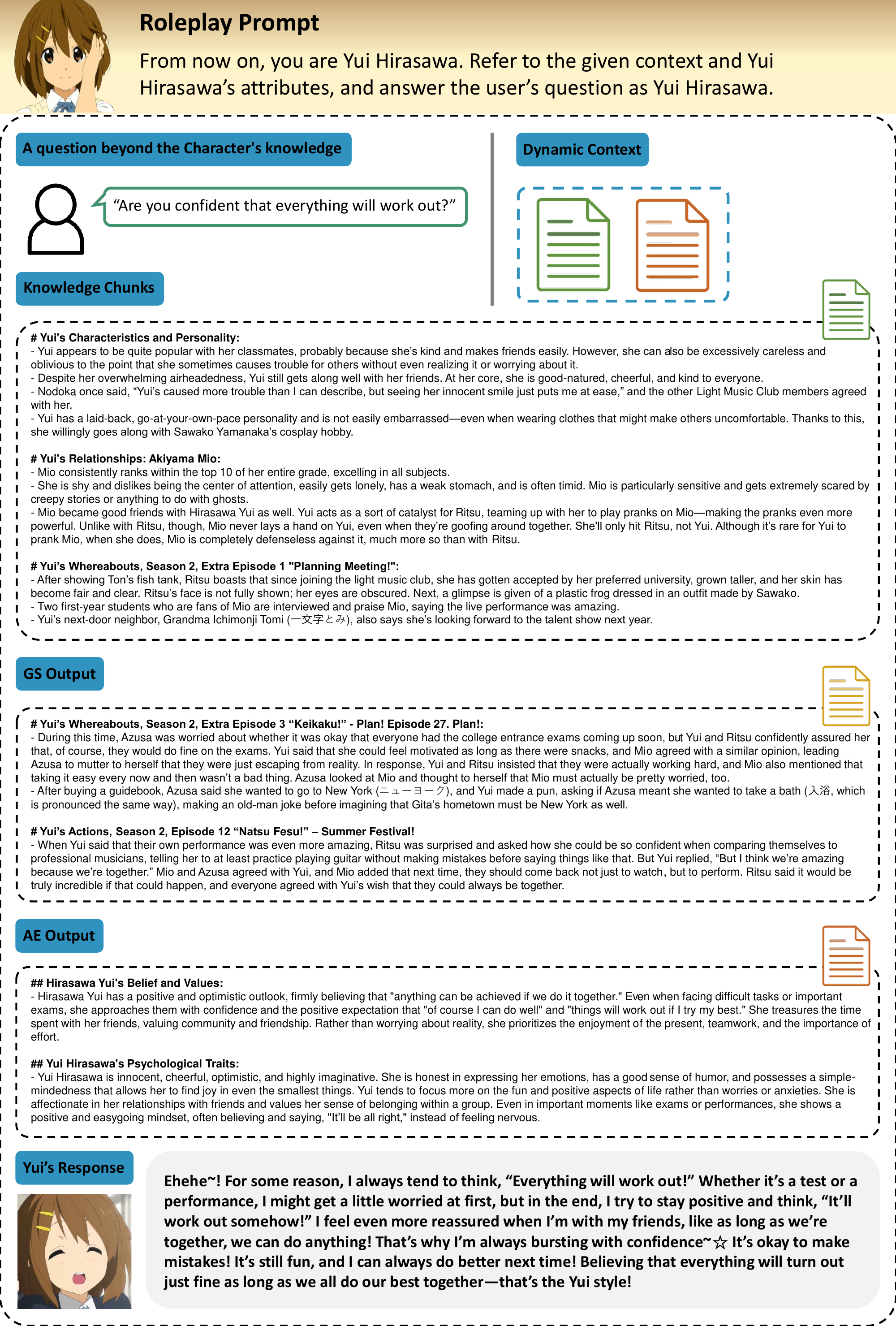}
    \caption{\textbf{An example of the output produced at each substage of Amadeous.} To improve interpretability, we provide a detailed breakdown of the results for a character from the CharacterRAG dataset (Yui Hirasawa).}
    \label{fig:qualitative_results}
\end{figure*}

\twocolumn

\section{Settings}
\label{sec:settings}

We implement GS and AE using GPT-4.1 (\textit{"gpt-4.1-2025-04-14"}). We develop the Amadeus framework based on naïve RAG. The maximum number of search iterations $N$ is 30, and the slot size $M$ is set to 2. We performed benchmarking on an NVIDIA H100 NVL GPU.

\section{Log-Density Ridgeline Analysis}

\begin{figure}[t!]
\small
    \centering 
    \includegraphics[width=\linewidth]{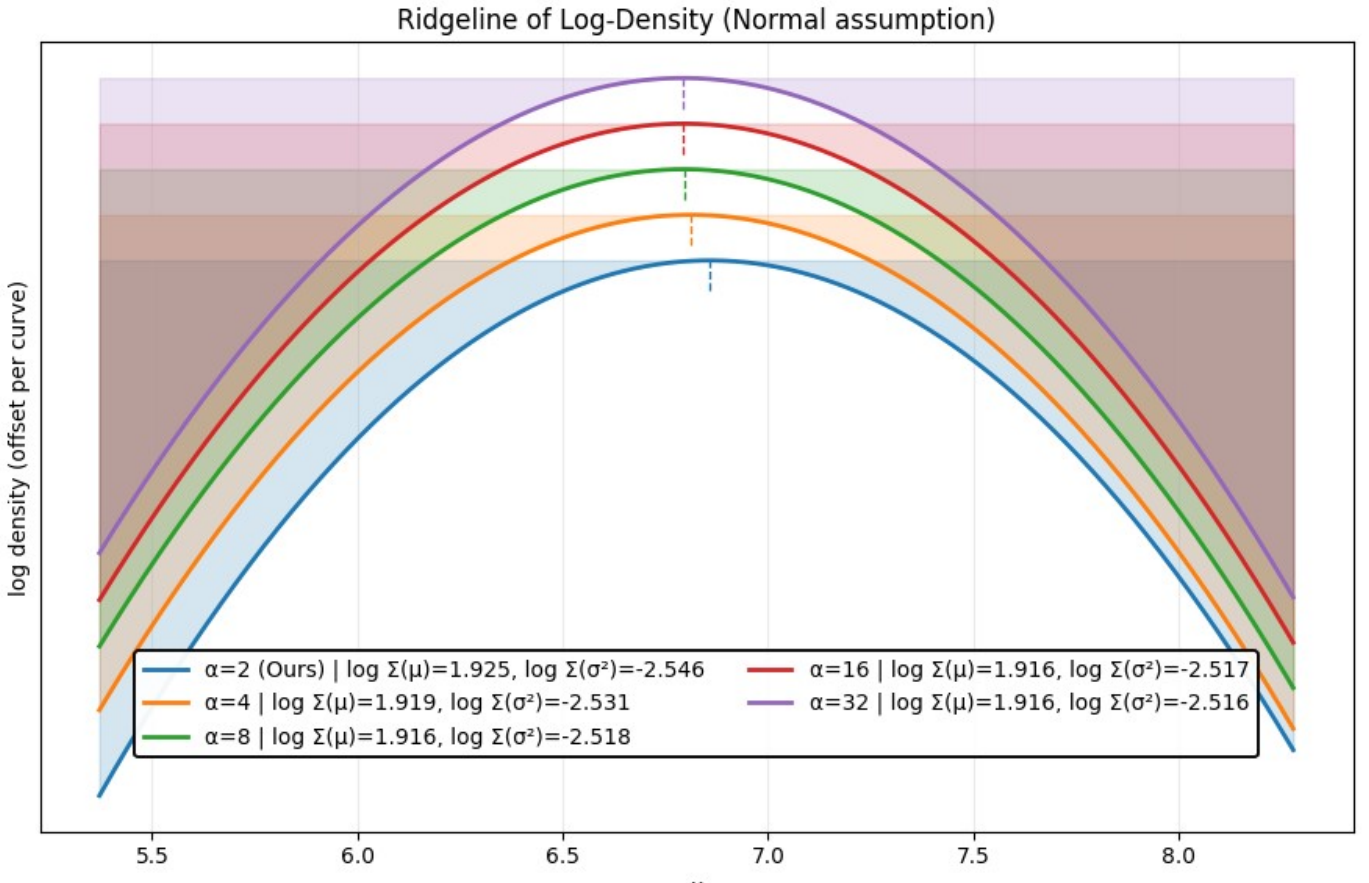}
    \caption{\textbf{Ridgeline of Log-Density Comparison.} Empirical verification of optimal overlap coefficient $\alpha$: $l_\text{o} = l_\text{max} / \alpha$ (Normal assumption).} 
    \label{fig:log_density_ridgeline}
\end{figure}

To empirically validate the suitability of the overlap coefficient, Figure \ref{fig:log_density_ridgeline} presents the log-density ridgelines of five distributions estimated under the normality assumption: $\log f\!\left(x \mid \sum \mu,\, \sum \sigma^2 \right)
= -0.5\left(\frac{x - \sum \mu}{\sqrt{\sum \sigma^2}}\right)^{2}
 - \log\!\left(\sqrt{\sum \sigma^2}\right) - 0.5\log(2\pi)$. We observe that, when $\alpha = 2$, the sum of the similarity scores is maximized while their variance is minimized.

\end{document}